\documentclass[sigconf]{acmart}
\acmSubmissionID{150}
\AtBeginDocument{%
  \providecommand\BibTeX{{%
    \normalfont B\kern-0.5em{\scshape i\kern-0.25em b}\kern-0.8em\TeX}}}

\copyrightyear{2023} 
\acmYear{2023} 
\setcopyright{acmlicensed}\acmConference[SA Technical Communications
'23]{SIGGRAPH Asia 2023 Technical Communications}{December 12--15,
2023}{Sydney, NSW, Australia}
\acmBooktitle{SIGGRAPH Asia 2023 Technical Communications (SA Technical
Communications '23), December 12--15, 2023, Sydney, NSW, Australia}
\acmPrice{15.00}
\acmDOI{10.1145/3610543.3626176}
\acmISBN{979-8-4007-0314-0/23/12}

\begin{document}
\title{Story-to-Motion: Synthesizing Infinite and Controllable Character Animation from Long Text}

\author{Zhongfei Qing}
\affiliation{%
  \institution{SenseTime Research}
  \city{BeiJing}
  \country{China}
  }
\email{qingzhongfei@sensetime.com}

\author{Zhongang Cai}
\affiliation{%
  \institution{SenseTime Research}
  \city{Singapore}
  \country{Singapore}
  }
\email{caizhongang@sensetime.com}

\author{Zhitao Yang}
\affiliation{%
  \institution{SenseTime Research}
  \city{ShenZhen}
  \country{China}
  }
\email{yangzhitao@sensetime.com}

\author{Lei Yang}
\authornote{corresponding author}
\affiliation{%
  \institution{SenseTime Research}
  \city{ShenZhen}
  \country{China}
  }
\email{yanglei@sensetime.com}

\newcommand{\misscite}{\textcolor{red}{[C]~}}
\newcommand{\missref}{\textcolor{red}{[R]~}}
\newcommand{\missvalue}{\textcolor{red}{[V]~}}
\newcommand{\needcheck}[1]{\textcolor{red}{#1}}
\newcommand{\todo}[1]{\textcolor{red}{#1}}



\begin{abstract}

  Generating natural human motion from a story has the potential to transform the landscape of animation, gaming, and film industries. 
  A new and challenging task, \textbf{Story-to-Motion}, arises when characters are required to move to various locations and perform specific motions based on a long text description.  This task demands a fusion of low-level control (trajectories) and high-level control (motion semantics). 
  Previous works in character control and text-to-motion have addressed related aspects, yet a comprehensive solution remains elusive: character control methods do not handle text description, whereas text-to-motion methods lack position constraints and often produce unstable motions.
  In light of these limitations, we propose a novel system that generates controllable, infinitely long motions and trajectories aligned with the input text.
  \textbf{1)} We leverage contemporary Large Language Models to act as a text-driven motion scheduler to extract a series of (text, position, duration) pairs from long text. 
  \textbf{2)} We develop a text-driven motion retrieval scheme that incorporates motion matching with motion semantic and trajectory constraints. 
  \textbf{3)} We design a progressive mask transformer that addresses common artifacts in the transition motion such as unnatural pose and foot sliding.
  Beyond its pioneering role as the first comprehensive solution for Story-to-Motion, our system undergoes evaluation across three distinct sub-tasks: trajectory following, temporal action composition, and motion blending, where it outperforms previous state-of-the-art (SOTA) motion synthesis methods across the board. Homepage: {\color{magenta}\url{https://story2motion.github.io/}}

\end{abstract}

\begin{teaserfigure}
  \includegraphics[width=\textwidth]{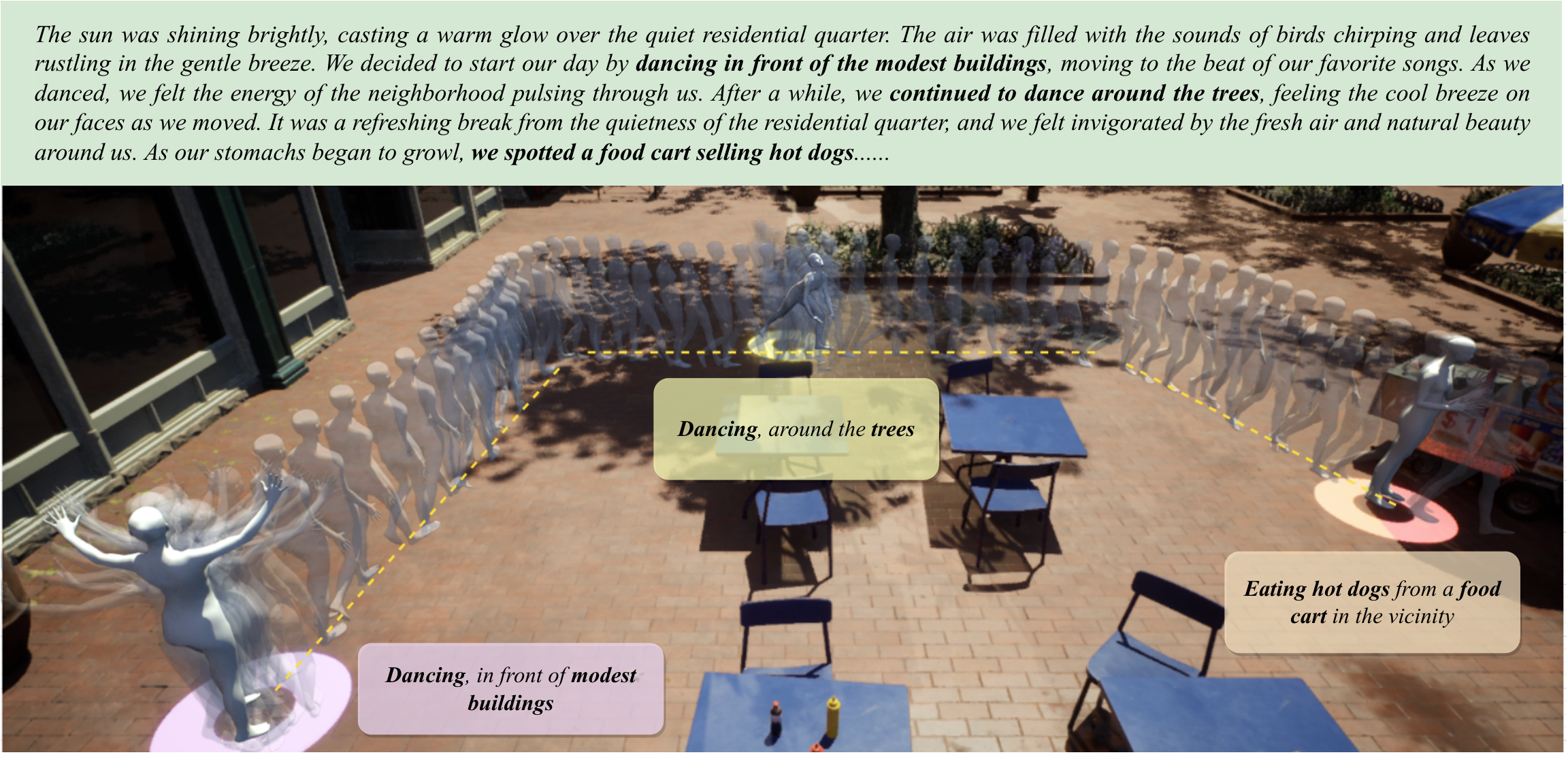}
  %\vspace{-7mm}
  %\vspace{-8mm}
  \caption{\emph{Story-to-Motion} is a new task that takes a story (top green area) and generates motions and trajectories that align with the text description.}
  \Description{Synthetic dataset.}
  \label{fig:teaser}
\end{teaserfigure}

\maketitle

\begin{figure*}[t]
  \centering
  %\vspace{-3mm}
  \includegraphics[width=\linewidth]{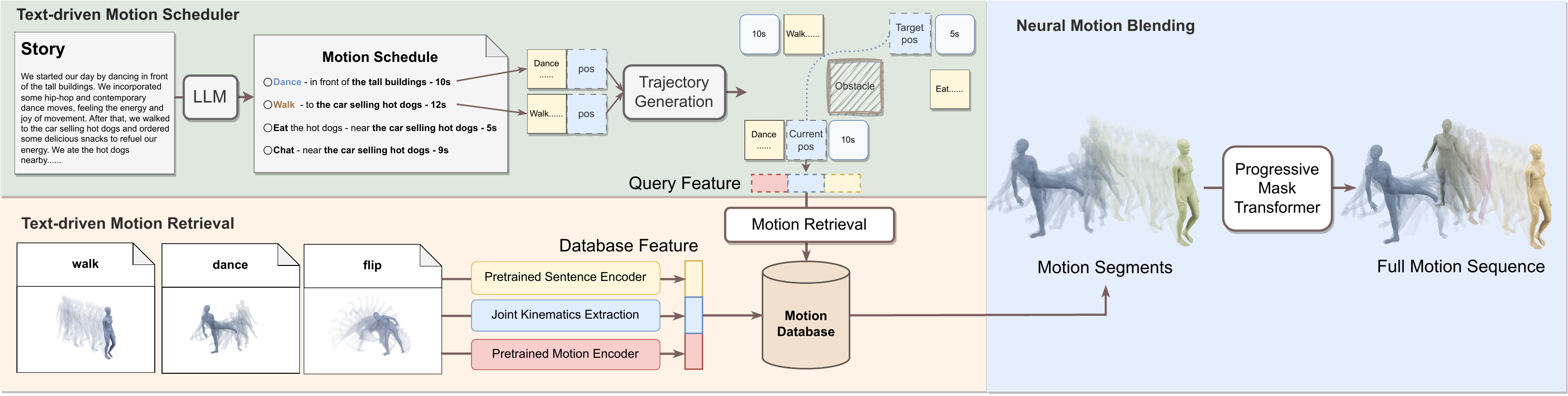}
  
  %\vspace{-3mm}
  \caption{The proposed \emph{Text-based Motion Matching} framework comprises three modules: (1) The Text-Driven Motion Scheduler extracts semantic information from the input story using a pretrained Large Language Model and obtains the trajectory based on known scene position information. (2) The Text-based Motion Retrieval module retrieves motions from the database that conform to both semantic and trajectory constraints. (3) Finally, the Neural Motion Blending module generates transition motion and concatenates motion segments into a natural-looking sequence.}
  \Description{The overview of the proposed method.}
  \label{fig:system}
  % \vspace{-3mm}
\end{figure*}

\section{Introduction}

%% Problem Definition

Imagine the potential of effortlessly translating a long textual narrative detailing a series of human activities traversing diverse locations into seamless, lifelike human motions. This transformation not only ushers in a new era of content generation but also has the power to reshape the animation, gaming, and film industries. Leveraging the capabilities of Large Language Models, this ambitious vision has now evolved from a distant goal into a tangible reality. Herein, we introduce a pioneering effort, \emph{Story-to-Motion}, a task that takes a "story" (exemplified in Fig.~\ref{fig:teaser}) as input and faithfully generates a sequence of motions that simultaneously conforms to low-level kinematic and high-level semantic constraints.

%% Existing solutions

%Existing methods primarily fall into two categories.
Previous works in character control and text-to-motion have addressed related aspects. 
Some efforts \cite{zhang2022gamma, holden2017phase} mainly focus on matching trajectories, supporting relatively few semantic motions, while the others \cite{zhang2023remodiffuse, zhang2022motiondiffuse} only focus on generating short semantic motions and ignore long motions with trajectories.
%Neither approach fully considers the combination of trajectory and semantic description
Neither approach fully integrates trajectory and semantic description, making it difficult to solve the problem effectively.
%Furthermore, deep learning-based methods are usually limited by the quality of the generated motion, especially for training sets with long-tail distribution of motions.
%For less frequent motions, such as handstands, they are easily ignored during training due to data imbalance, resulting in unsatisfactory results during inference.
%To tackle the story-to-motion problem and achieve satisfactory motion quality, we propose a novel \emph{Text-based Motion Matching} framework.
Besides, directly generating motions with learning-based methods is limited by motion quality, particularly for long-tail training sets. 
%Less frequent motions like handstands are disregarded due to data imbalance, leading to unsatisfactory outcomes.
Neglected due to data imbalance, less frequent motions like handstands yield unsatisfactory results.

%% Key design of our method
To generate high-quality and long motion, motion matching is widely adopted in the industry. 
While practical, it cannot be directly applied to this problem due to its inability to utilize text input and the tendency for blending artifacts.
To address these challenges, we incorporate text embedding to match candidate motions. Furthermore, kinematics features and learned features are leveraged for further retrieval. Additionally, dynamic target trajectory is proposed to improve trajectory matching. 
Besides, heuristic-based blending algorithms yield subpar results with complex motions. To address this, we design a progressive mask transformer for motion transitions.

%% Contributions

Our contributions are three-fold:
1) We propose a new task, \emph{Story-to-Motion}, which considers both trajectory and semantics when generating motions.
2) We propose \emph{Text-based Motion Matching}, a promising long text-driven, controllable system to address this task.
3) %We experiment with three sub-tasks using standard datasets for comparison, which show that our design outperforms the current SOTA in all three tasks. 
Through experimentation on standard datasets, our design outperforms the current SOTA methods in all three sub-tasks.
\section{Related Works}

%\noindent\textbf{Trajectory-based Motion Synthesis.} Generating motion that follows a given trajectory has garnered significant attention. Traditional algorithms employ motion matching \cite{clavet2016motion}, a renowned retrieval-based motion control technique. This approach involves matching the current character pose with segments of animation stored in a database, based on the target trajectory. Learned motion matching \cite{holden2020learned} employs an auto-regressive neural network to predict the next motion state based on a given control signal. Moreover, Phase-Functioned Neural Networks \cite{holden2017phase} incorporates the phase as input and calculate the weights using a cyclic function.
%\noindent\textbf{Trajectory-based Motion Synthesis.} 
\subsection{Trajectory-based Motion Synthesis}
To generate motion from a given trajectory, motion matching \cite{clavet2016motion} retrieves segments of animation stored in a database, based on the current pose and the target trajectory. Learned motion matching \cite{holden2020learned} employs an auto-regressive neural network to predict the next motion state based on a given control signal. Moreover, Phase-Functioned Neural Networks \cite{holden2017phase} incorporates a cyclic Phase Function to generate the network weights. GAMMA \cite{zhang2022gamma} employs reinforcement learning to generative motion primitives via body surface markers.

%\noindent\textbf{Text-based Motion Synthesis.}
\subsection{Text-based Motion Synthesis}
Earlier research in motion synthesis from text such as JL2P~\cite{ahuja2019language2pose} constructs a joint space to which both text descriptions and motion sequences are mapped. 
Variational mechanisms have also been introduced for higher diversity. For example, TEACH~\cite{athanasiou2022teach} utilizes transformer-based VAEs to produce motion with text conditions and is able to achieve temporal action compositionality.
Moreover, MDM~\cite{tevet2022humanmotiondiffusion}, MotionDiffuse~\cite{zhang2022motiondiffuse} adapt diffusion models to generate human motions from text input. ReMoDiffuse~\cite{zhang2023remodiffuse} further integrates a retrieval mechanism to refine the denoising process and enhance the generalizability and diversity.

%\noindent\textbf{Motion blending.}
\subsection{Motion blending}
Motion blending can be considered as a specialized form of motion completion \cite{duan2021single}, where the resulting motion is constrained by given context frames. Traditional solutions involve interpolating keyframes using techniques such as Bezier curves or polynomials \cite{bollo2016inertialization}. In the realm of deep learning methods, SOTA approaches primarily rely on transformer networks. Duan et al. \cite{duan2021single} utilize a pretrained language model encoder and 1D convolution to generate transition motion. Qin et al. \cite{qin2022twostage} propose a two-stage approach that demonstrates improved ability for longer in-betweenings by utilizing Keyframe Positional Encoding and Learned Relative Positional Encoding. 
\section{Methodology}

\subsection{Text-driven Motion Scheduler}
%After a textual story is given, 
Given a story, the Text-based Motion Scheduler module prompts ChatGPT~\cite{chatGPT} in natural language for a list of text descriptions of the character's actions $\mathcal{T}=\{text_i\}$, location names $\mathcal{L}$, and duration $t$ of those actions (Fig.~\ref{fig:system}). 
Assuming that the 3D scene is known, we can look up the corresponding coordinates $\mathcal{P} = \{(x_i,y_i)\}$ with $\mathcal{L}$.
Then, the trajectory generation module converts the discrete $\mathcal{P}$ into a continuous curve via a path-finding algorithm.
%\cite{sharon2015conflict}.
%
For discrete textual descriptions, each motion is additionally given a designated duration by LLM, while the remaining idle time is filled in with ``walking''. Thus, we transform the story into a continuous function over time, named \emph{Scheduler} $\mathcal{S}(t)$, where each time point $t$ corresponds to $(x_i, y_i, text_i)$, containing both low-level locations $(x_i, y_i)$ and high-level textual descriptions $text_i$ about motion.

\begin{figure}[t]
  \centering
  \includegraphics[width=\linewidth]{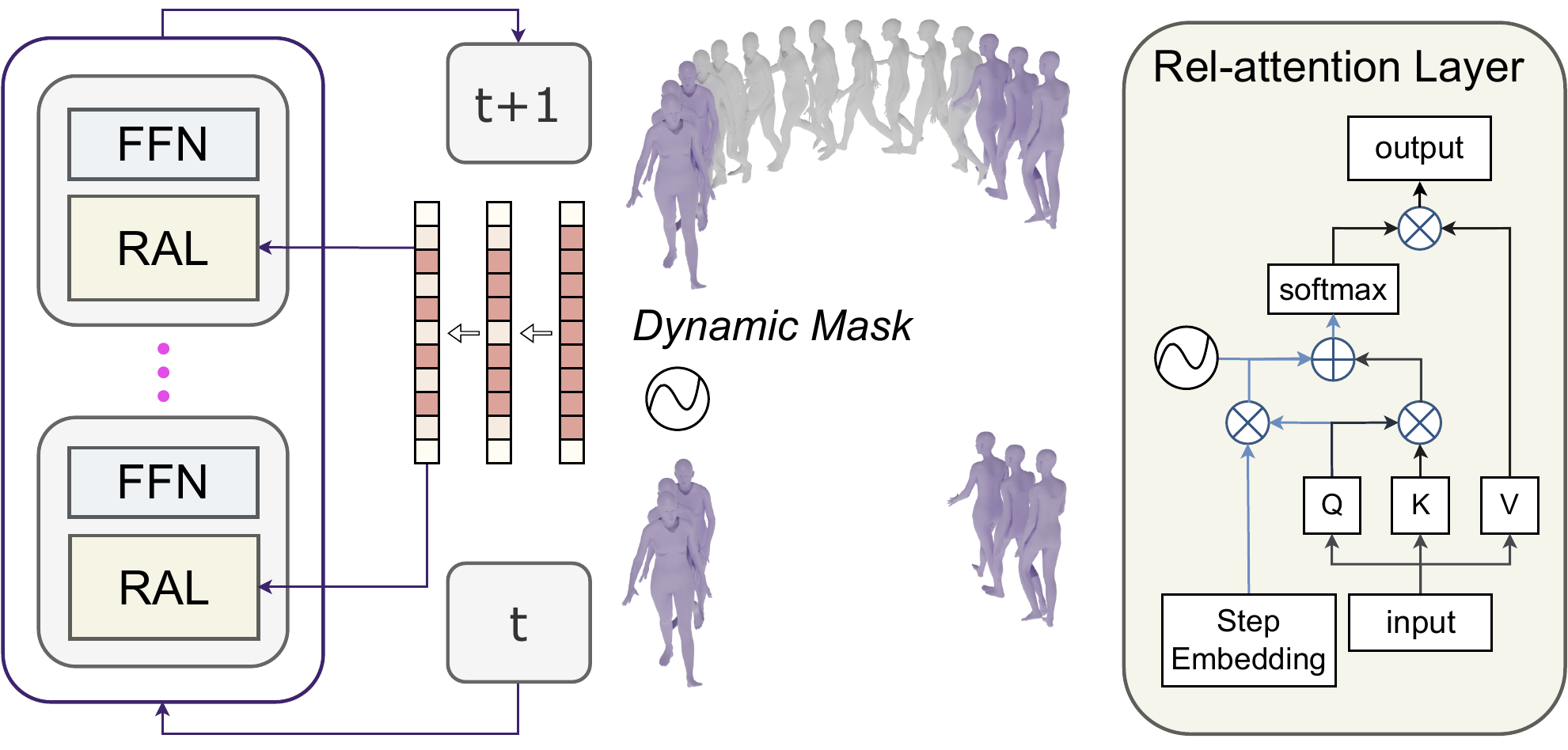}
  %\vspace{-5mm}
  \caption{
  The Progressive Mask Transformer generates motion over multiple iterations (e.g., 3), with the number of masked missing frames decreasing linearly. FFN: feedforward network. RAL: Rel-attention Layer.
  }
  \Description{User study.}
  \label{fig:progressive}
%\vspace{-3mm}
\end{figure}

\subsection{Text-based Motion Retrieval}

Given a motion database $\mathcal{D} = \{(text_i, m_i)\}$, the \emph{Text-based Motion Retrieval} module matches motions in an auto-regressive manner. 
Here, $text_i$ is the text label (e.g., "sitting") 
while $m_i \in \mathbb{R}^{L \times D}$ denotes a motion sequence with $L$ frames. Each frame is a $D$-dimensional vector representing body joint rotation and overall translation.
It is notable that the database consists of short clips. The duration of the target action is included in ChatGPT-generated instruction, which affects the number of clips. Hence, the system is scalable to arbitrary lengths of motion, as the main idea is to retrieve short clips and blend them into long motions.

At $t=0$, a motion $m_0$ is randomly selected based on $text_0$ and placed at $(x_0, y_0)$. At the time point $t$, given $(x_{t}, y_{t}, text_{t})$ from $\mathcal{S}(t)$ and the previous motion $m_{t-1}$, metrics are used to find the next best matching motion from the database.
The objective is to find a motion clip that aligns with both the query text and trajectory, while also maintaining a similar body pose and motion style as the previous motion for coherence. We achieve this goal in two steps. 

In the first step, to incorporate semantic information, a pre-trained sentence encoder \cite{liu2019roberta} extracts text embedding $f_{text}$ from the given text. 
%Top-$K_1$ candidates are selected via cosine similarity for subsequent matching. 
Top-$K_1$ results, selected via cosine similarity, serve as candidates for subsequent matching.
However, two challenges arise: 
(1) Imperfect text matching results in irrelevant motions.
(2) Some datasets contain a small amount of low-quality clips.
Consequently, the presence of noisy matched results will adversely affect the quality of the generated motion.
Thus outlier removal is employed to reject noisy motion clips that are far from the distribution center.

In the second step, trajectory and coherence constraints are incorporated through motion-matching. The crucial aspect is determining the similarity measure. 
The original Motion Matching~\cite{holden2020learned} method mainly adopts trajectory and joint position similarity. The features they used include lower body part $f_{lower} = \{ foot, \  \dot{foot}, \  \dot{hip} \} $ and trajectory part $f_{traj} = \{pos, \  direction\}$, where $pos$ denote the 2D future trajectory positions projected on the ground, $direction$ are the future trajectory facing directions, $foot$ are the two foot joint positions, $\dot{foot}$ are the two foot joint velocities, and $\dot{hip}$ is the hip joint velocity.
However, this strategy overlooks the consistency of the upper-body, resulting in potential swaying of the upper-body. Moreover, sudden changes in motion style can negatively impact the visual quality. %The motion coherence is important for long-term motion that aligns with long text.
Therefore, we include the upper-body feature $f_{upper} = \{ upper-body, \  \dot{upper-body}\} $ which is the upper-body joints positions and velocities. Moreover, we train an auto-encoder model $\mathcal{F}$ to extract motion features $f_{learned}$ that encompass the entire body information, together with temporal cues and motion style, enhancing the matching capability of hand-crafted features. 
With all designed features, we compute the Euclidean similarity for each of them, including lower-body, upper-body, trajectory, and learned features. Z-Score normalizing the features is crucial due to their potential significant differences in magnitudes.
To summarize, we first use text embeddings to select top-$K_1$ candidate motions, 
then the top-$K_2$ desired motions are selected by the weighted sum $S$ of the above similarity, with adjustable weights for low-level and high-level scenarios. To ensure diversity, we choose clips (e.g., 10) randomly from the most suitable candidates.

\begin{table*}[t] 
    %\vspace{-1mm}
    \caption{Motion Blending Benchmark in AMASS, all models are trained with random transition lengths from 5 to 60. The numbers in red and blue indicate the best and the second-best results. Our progressive mask transformer surpasses previous state-of-the-art motion completion methods.}
    \label{tab:amassExp}
    %\vspace{-3mm}
    \centering
    \resizebox{\textwidth}{!}{
    \begin{tabular}{lllllllllllllll}
    \toprule
        ~ & Param & \multicolumn{6}{c}{Pos / m} & \multicolumn{6}{c}{Rotation} & Mean \\ 
        Frames & ~ & 5 & 15 & 30 & 45 & 60 & 70 & 5 & 15 & 30 & 45 & 60 & 70  \\ 
    \midrule
FCN&15.89M&0.68 &0.812 &1.113 &1.524 &2.157 &2.71 &0.585 &0.648 &0.795 &0.932 &1.098 &1.292 &1.1955 \\
MC-Trans \cite{duan2021single}&13.03M&0.362 &0.667 &0.993 &1.308 &1.584 &1.872 &0.198 &0.375 &0.578 &0.720 &0.829 &0.921 &0.867 \\
Transformer \cite{vaswani2017attention} &10.12M&0.258 &0.522 &0.865 &1.198 &1.47 &1.703 &0.212 &0.352 &0.542 &0.684 &0.794 &0.885 &0.7904 \\
Context Trans \cite{qin2022twostage}&10.39M&0.243 &0.488 &0.829 &1.145 &1.416 &1.637 &0.195 &0.321 &0.509 &0.662 &0.778 &0.863 &0.757 \\

Detail Trans \cite{qin2022twostage}&20.52M&\color{blue}{0.142} &\color{blue}{0.398} &0.762 &1.082 &1.358 &1.576 &\color{blue}{0.123} &\color{blue}{0.265} &\color{blue}{0.465} &\color{blue}{0.624} &0.744 &0.828 &0.697 \\
%Ours (w/o progressive gen)&10.39M&0.224 &0.459 &0.806 &1.129 &1.396 &1.63 &0.177 &0.305 &0.499 &0.654 &0.762 &0.846 &0.7406 \\
Ours (one stage)&10.39M&0.167 &0.402 &\color{blue}{0.728} &\color{blue}{1.030} &\color{blue}{1.296} &\color{red}{1.506} &0.161 &0.290 &0.474 &0.636 &\color{blue}{0.735} &\color{blue}{0.819} &\color{blue}{0.687} \\
Ours (two stages)&20.77M&\color{red}{0.122} &\color{red}{0.379} &\color{red}{0.714} &\color{red}{1.022} &\color{red}{1.291} &\color{blue}{1.508} &\color{red}{0.113} &\color{red}{0.262} &\color{red}{0.455} &\color{red}{0.621} &\color{red}{0.726} &\color{red}{0.817} &\color{red}{0.669} \\
    \bottomrule
    \end{tabular}
}
%\vspace{-2mm}
\end{table*}

Iterating the aforementioned matching process can generate arbitrarily long motions. However, its auto-regressive nature can lead to cumulative trajectory errors. To address this, the target trajectory is dynamically adjusted based on the current position and target position: when a position error occurs, the subsequent step will correct this error by retrieving clips that minimize the error.

\subsection{Neural Motion Blending}
\label{mask}
Motion matching generates a sequence of motion clips, which can be quite numerous when dealing with long text. Thus realistic transition motions is the key to high-quality result. 
The current SOTA method two-stage transformer~\cite{qin2022twostage} has two limitations:
(1) It employs a full zero attention mask in the first stage, which results in unreliable information propagation when the length of the mask is large.
(2) Although the two-stage design shows promising results, their parameters are not shared, which not only causes inefficiency but also prevents the second stage from fully utilizing the information learned in the first stage.
We design a \emph{Progressive Mask Transformer} (Fig.\ref{fig:progressive}) to tackle these problems. Inspired by the progressive strategy used in other generation tasks~\cite{maskpredict}, we propose a coarse-to-fine approach that generates motion progressively, sharing parameters among iterative steps. This approach is both parameter-efficient and can leverage knowledge from a previously well-learned model.
%
%Particularly, to handle the issue of missing information in long motion, we design a dynamic attention mask that gradually introduces more information in each iteration, ensuring reliable information propagation. 
Particularly, to ensure reliable information propagation, we design a dynamic attention mask that gradually introduces more information in each iteration. 
Refer to Appendix A for details.

\section{EXPERIMENTS}

To our best knowledge, no prior work can generate infinitely long motions and trajectories aligned with the given long text. Hence we compare our method with SOTA techniques in three sub-tasks: trajectory following, temporal action composition, and motion blending, to gauge different aspects of our system. 
Supplementary materials include an overall visualization. 
The experiments are conducted on the database AMASS \cite{mahmood2019amass} that unifies different datasets.
More details are in Appendix B. 

\subsection{Trajectory Following}

We compare our system with GAMMA \cite{zhang2022gamma}, the current SOTA method for infinite long motion generation, and closely follow their experiment settings.
As indicated in Table \ref{tab:trajectory}, it shows significant advantages in trajectory following (columns 4-6). GAMMA often takes a long time to reach a nearby goal with sudden speed changes or stops \cite{zhang2022gamma}. In contrast, thanks to the retrieval strategy that takes into account both speed and position, our method can faithfully follow the trajectory and maintain control over speed, which is crucial for generating multi-character motion and avoiding collisions. 
%Additionally, addressing penetration and floating errors is challenging, where our retrieval-based system exhibits significant advantages (columns 1-3).
Moreover, our retrieval-based system excels in addressing penetration and floating errors (columns 1-3).

\begin{figure}[t!]
  \centering
  \includegraphics[width=\linewidth]{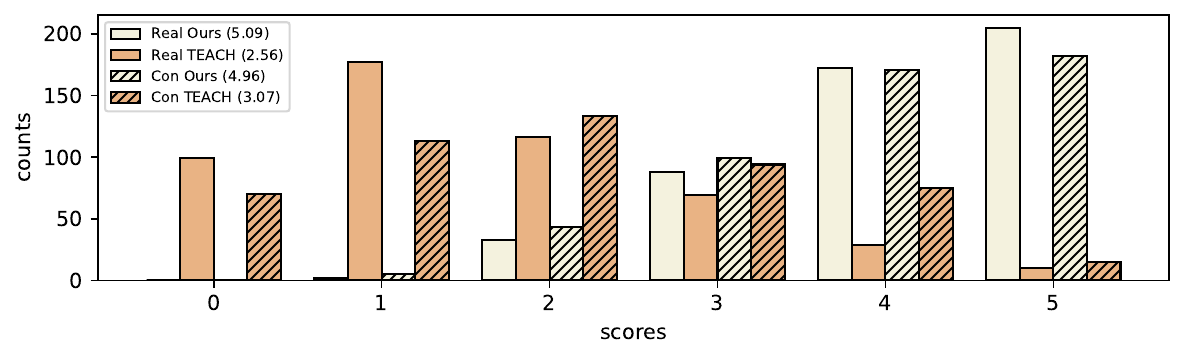}
  %\vspace{-8mm}
  \caption{
  A user study on temporal action composition (number in parentheses indicates average score). We selected 25 subjects (6 women and 19 men), with ages ranging from 20 to 35, including animators, AI researchers, and gaming enthusiasts. Our method is deemed more realistic and text-consistent.
  }
  \Description{User study.}
  \label{fig:user}
  %\vspace{-3mm}
\end{figure}

\begin{table}[t] 
    \caption{
    Long motion generation with trajectory and speed constraints. Our method shows stronger trajectory following ability with less physics error.
    }
    \label{tab:trajectory}
    \centering
    %\vspace{-3mm}
    \begin{tabular}{lllllll}
    \toprule
        ~  & \multicolumn{3}{c}{Physics Error / m} & \multicolumn{3}{c}{Trajectory Error / m} \\ 
        Trajectory & Wave  & Circle & Square  & Wave  & Circle & Square   \\ 
    \midrule

GAMMA & 0.050 &0.047 &0.058 &2.306 &1.951 &3.242 \\
Ours & \color{red}{0.025} & \color{red}{0.030} &\color{red}{0.020} & \color{red}{0.156} & \color{red}{0.249} &\color{red}{0.151} \\
    \bottomrule
    \end{tabular}
    %\vspace{-3mm}
\end{table}

\subsection{Temporal Action Composition}

%The temporal action composition is to generate motions that correspond to a series of text descriptions and follow the temporal order of the instructions. 
Temporal action composition is to generate motions aligned with a series of text descriptions, following the prescribed temporal order.
%We conduct a user study considering text consistency is hard to evaluate \cite{holden2020learned}. We compared our method with the current SOTA method TEACH \cite{athanasiou2022teach}. 
We conduct a user study to compare our method with the current SOTA method TEACH \cite{athanasiou2022teach}, considering text consistency is hard to evaluate \cite{holden2020learned}.
%Temporal action composition is to generate motions from a series of text descriptions, following their temporal order. We compared our method with the current SOTA method TEACH \cite{athanasiou2022teach} by user study, considering text consistency is hard to evaluate \cite{holden2020learned}. 
As shown in Figure \ref{fig:user}, the proposed system shows superior performance in both realism and text-consistency scores, with much fewer artifacts like floating and penetration. 
%Besides, it is extremely challenging for deep-learning methods to generate complicated and rare motions. 
Besides, deep-learning methods face significant challenges in generating rare motions.
Our system can produce great quality in this situation while TEACH fails.

\subsection{Motion Blending}
%In the motion blending task, all the models undergo comprehensive evaluation across transitions of 5 to 70 frames.
In the motion blending task, we extensively evaluate models on transitions spanning from 5 to 70 frames. 
As shown in Table \ref{tab:amassExp}, our system shows superior performance compared with previous works. It exhibits great improvement (an average 37\% relative improvement) compared with MC-Trans \cite{duan2021single} (rows 2 and 6) which performs well in LAFAN1. Besides, the proposed method brings a considerable improvement (an average 15\% relative improvement) compared with the strong baseline \cite{qin2022twostage} (rows 4 and 6). 
Notably, our single-stage method surpasses the previous SOTA two-stage transformers \cite{qin2022twostage} (rows 5 and 6) using nearly half of the parameters, which implies that the masked attention it used in the first stage may not be suitable for long-term motion generation since it cuts off the connection between missing frames. Using the two-stage approach \cite{qin2022twostage}, our method demonstrates additional advancement (row 7).

\section{Conclusion}

In this work, we propose a new task, \emph{Story-to-Motion}, with the goal of generating human motion and trajectory from a long text. Moreover, we present a pioneering effort, \emph{Text-based Motion Matching}, that leverages a large language model, motion matching, and neural blending for controllable and realistic motion generation. 
It surpasses previous SOTA methods in three sub-tasks: trajectory following, temporal action composition, and motion blending.

%\noindent \textbf{Limitation.} 
Since our system is retrieval-based, its diversity is ultimately limited by the size of the motion database. Thus it is promising to combine our method with learning-based methods. %Thus it is promising to combine our method with learning-based methods.

\bibliographystyle{ACM-Reference-Format}
\bibliography{reference}


\begin{thebibliography}{35}

%%% the Web address.

\ifx \showCODEN    \undefined \def \showCODEN     #1{\unskip}     \fi
\ifx \showDOI      \undefined \def \showDOI       #1{#1}\fi
\ifx \showISBNx    \undefined \def \showISBNx     #1{\unskip}     \fi
\ifx \showISBNxiii \undefined \def \showISBNxiii  #1{\unskip}     \fi
\ifx \showISSN     \undefined \def \showISSN      #1{\unskip}     \fi
\ifx \showLCCN     \undefined \def \showLCCN      #1{\unskip}     \fi
\ifx \shownote     \undefined \def \shownote      #1{#1}          \fi
\ifx \showarticletitle \undefined \def \showarticletitle #1{#1}   \fi
\ifx \showURL      \undefined \def \showURL       {\relax}        \fi
% invisible to TeX
\providecommand\bibfield[2]{#2}
\providecommand\bibinfo[2]{#2}
\providecommand\natexlab[1]{#1}
\providecommand\showeprint[2][]{arXiv:#2}

\bibitem[{Advanced Computing Center for the Arts and Design}({[n.\,d.]})]%
        {ACCAD}
\bibfield{author}{\bibinfo{person}{{Advanced Computing Center for the Arts and Design}}.} \bibinfo{year}{[n.\,d.]}\natexlab{}.
\newblock \bibinfo{title}{{ACCAD MoCap Dataset}}.
\newblock
\newblock
\urldef\tempurl%
\url{https://accad.osu.edu/research/motion-lab/mocap-system-and-data}
\showURL{%
\tempurl}


\bibitem[Ahuja and Morency(2019)]%
        {ahuja2019language2pose}
\bibfield{author}{\bibinfo{person}{Chaitanya Ahuja} {and} \bibinfo{person}{Louis-Philippe Morency}.} \bibinfo{year}{2019}\natexlab{}.
\newblock \showarticletitle{Language2pose: Natural language grounded pose forecasting}. In \bibinfo{booktitle}{\emph{3DV}}. IEEE, \bibinfo{pages}{719--728}.
\newblock


\bibitem[Aristidou et~al\mbox{.}(2019)]%
        {DanceDB}
\bibfield{author}{\bibinfo{person}{Andreas Aristidou}, \bibinfo{person}{Ariel Shamir}, {and} \bibinfo{person}{Yiorgos Chrysanthou}.} \bibinfo{year}{2019}\natexlab{}.
\newblock \showarticletitle{Digital Dance Ethnography: {O}rganizing Large Dance Collections}.
\newblock \bibinfo{journal}{\emph{J. Comput. Cult. Herit.}} \bibinfo{volume}{12}, \bibinfo{number}{4}, Article \bibinfo{articleno}{29} (\bibinfo{date}{Nov.} \bibinfo{year}{2019}), \bibinfo{numpages}{27}~pages.
\newblock
\showISSN{1556-4673}
\urldef\tempurl%
\url{https://doi.org/10.1145/3344383}
\showDOI{\tempurl}


\bibitem[Athanasiou et~al\mbox{.}(2022)]%
        {athanasiou2022teach}
\bibfield{author}{\bibinfo{person}{Nikos Athanasiou}, \bibinfo{person}{Mathis Petrovich}, \bibinfo{person}{Michael~J Black}, {and} \bibinfo{person}{G{\"u}l Varol}.} \bibinfo{year}{2022}\natexlab{}.
\newblock \showarticletitle{TEACH: Temporal Action Composition for 3D Humans}.
\newblock \bibinfo{journal}{\emph{arXiv:2209.04066}} (\bibinfo{year}{2022}).
\newblock


\bibitem[Bollo(2016)]%
        {bollo2016inertialization}
\bibfield{author}{\bibinfo{person}{David Bollo}.} \bibinfo{year}{2016}\natexlab{}.
\newblock \showarticletitle{Inertialization: High-performance animation transitions in’gears of war’}.
\newblock \bibinfo{journal}{\emph{Proc. of GDC 2018}} (\bibinfo{year}{2016}).
\newblock


\bibitem[Brown et~al\mbox{.}(2020)]%
        {chatGPT}
\bibfield{author}{\bibinfo{person}{Tom Brown}, \bibinfo{person}{Benjamin Mann}, \bibinfo{person}{Nick Ryder}, \bibinfo{person}{Melanie Subbiah}, \bibinfo{person}{Jared~D Kaplan}, \bibinfo{person}{Prafulla Dhariwal}, \bibinfo{person}{Arvind Neelakantan}, \bibinfo{person}{Pranav Shyam}, \bibinfo{person}{Girish Sastry}, \bibinfo{person}{Amanda Askell}, {et~al\mbox{.}}} \bibinfo{year}{2020}\natexlab{}.
\newblock \showarticletitle{Language models are few-shot learners}.
\newblock \bibinfo{journal}{\emph{NeurIPS}}  \bibinfo{volume}{33} (\bibinfo{year}{2020}), \bibinfo{pages}{1877--1901}.
\newblock


\bibitem[{Carnegie Mellon University}({[n.\,d.]})]%
        {CMU}
\bibfield{author}{\bibinfo{person}{{Carnegie Mellon University}}.} \bibinfo{year}{[n.\,d.]}\natexlab{}.
\newblock \bibinfo{title}{{CMU MoCap Dataset}}.
\newblock
\newblock
\urldef\tempurl%
\url{http://mocap.cs.cmu.edu}
\showURL{%
\tempurl}


\bibitem[Clavet(2016)]%
        {clavet2016motion}
\bibfield{author}{\bibinfo{person}{Simon Clavet}.} \bibinfo{year}{2016}\natexlab{}.
\newblock \showarticletitle{Motion matching and the road to next-gen animation}. In \bibinfo{booktitle}{\emph{Proc. of GDC}}, Vol.~\bibinfo{volume}{2016}.
\newblock


\bibitem[Duan et~al\mbox{.}(2021)]%
        {duan2021single}
\bibfield{author}{\bibinfo{person}{Yinglin Duan}, \bibinfo{person}{Tianyang Shi}, \bibinfo{person}{Zhengxia Zou}, \bibinfo{person}{Yenan Lin}, \bibinfo{person}{Zhehui Qian}, \bibinfo{person}{Bohan Zhang}, {and} \bibinfo{person}{Yi Yuan}.} \bibinfo{year}{2021}\natexlab{}.
\newblock \showarticletitle{Single-shot motion completion with transformer}.
\newblock \bibinfo{journal}{\emph{arXiv:2103.00776}} (\bibinfo{year}{2021}).
\newblock


\bibitem[Ghazvininejad et~al\mbox{.}(2019)]%
        {maskpredict}
\bibfield{author}{\bibinfo{person}{Marjan Ghazvininejad}, \bibinfo{person}{Omer Levy}, \bibinfo{person}{Yinhan Liu}, {and} \bibinfo{person}{Luke Zettlemoyer}.} \bibinfo{year}{2019}\natexlab{}.
\newblock \showarticletitle{Mask-predict: Parallel decoding of conditional masked language models}.
\newblock \bibinfo{journal}{\emph{arXiv:1904.09324}} (\bibinfo{year}{2019}).
\newblock


\bibitem[Ghorbani et~al\mbox{.}(2021)]%
        {ghorbani2021movi}
\bibfield{author}{\bibinfo{person}{Saeed Ghorbani}, \bibinfo{person}{Kimia Mahdaviani}, \bibinfo{person}{Anne Thaler}, \bibinfo{person}{Konrad Kording}, \bibinfo{person}{Douglas~James Cook}, \bibinfo{person}{Gunnar Blohm}, {and} \bibinfo{person}{Nikolaus~F Troje}.} \bibinfo{year}{2021}\natexlab{}.
\newblock \showarticletitle{MoVi: A large multi-purpose human motion and video dataset}.
\newblock \bibinfo{journal}{\emph{Plos one}} \bibinfo{volume}{16}, \bibinfo{number}{6} (\bibinfo{year}{2021}), \bibinfo{pages}{e0253157}.
\newblock


\bibitem[Harvey et~al\mbox{.}(2020)]%
        {harvey2020robust}
\bibfield{author}{\bibinfo{person}{F{\'e}lix~G Harvey}, \bibinfo{person}{Mike Yurick}, \bibinfo{person}{Derek Nowrouzezahrai}, {and} \bibinfo{person}{Christopher Pal}.} \bibinfo{year}{2020}\natexlab{}.
\newblock \showarticletitle{Robust motion in-betweening}.
\newblock \bibinfo{journal}{\emph{ACM Transactions on Graphics (TOG)}} \bibinfo{volume}{39}, \bibinfo{number}{4} (\bibinfo{year}{2020}), \bibinfo{pages}{60--1}.
\newblock


\bibitem[Holden et~al\mbox{.}(2020)]%
        {holden2020learned}
\bibfield{author}{\bibinfo{person}{Daniel Holden}, \bibinfo{person}{Oussama Kanoun}, \bibinfo{person}{Maksym Perepichka}, {and} \bibinfo{person}{Tiberiu Popa}.} \bibinfo{year}{2020}\natexlab{}.
\newblock \showarticletitle{Learned motion matching}.
\newblock \bibinfo{journal}{\emph{ACM TOG}} \bibinfo{volume}{39}, \bibinfo{number}{4} (\bibinfo{year}{2020}), \bibinfo{pages}{53--1}.
\newblock


\bibitem[Holden et~al\mbox{.}(2017)]%
        {holden2017phase}
\bibfield{author}{\bibinfo{person}{Daniel Holden}, \bibinfo{person}{Taku Komura}, {and} \bibinfo{person}{Jun Saito}.} \bibinfo{year}{2017}\natexlab{}.
\newblock \showarticletitle{Phase-functioned neural networks for character control}.
\newblock \bibinfo{journal}{\emph{ACM TOG}} \bibinfo{volume}{36}, \bibinfo{number}{4} (\bibinfo{year}{2017}), \bibinfo{pages}{1--13}.
\newblock


\bibitem[Krebs et~al\mbox{.}(2021)]%
        {KrebsMeixner2021}
\bibfield{author}{\bibinfo{person}{Franziska Krebs}, \bibinfo{person}{Andre Meixner}, \bibinfo{person}{Isabel Patzer}, {and} \bibinfo{person}{Tamim Asfour}.} \bibinfo{year}{2021}\natexlab{}.
\newblock \showarticletitle{The KIT Bimanual Manipulation Dataset}. In \bibinfo{booktitle}{\emph{IEEE/RAS International Conference on Humanoid Robots (Humanoids)}}. \bibinfo{pages}{499--506}.
\newblock


\bibitem[Li et~al\mbox{.}(2022)]%
        {li2022mat}
\bibfield{author}{\bibinfo{person}{Wenbo Li}, \bibinfo{person}{Zhe Lin}, \bibinfo{person}{Kun Zhou}, \bibinfo{person}{Lu Qi}, \bibinfo{person}{Yi Wang}, {and} \bibinfo{person}{Jiaya Jia}.} \bibinfo{year}{2022}\natexlab{}.
\newblock \showarticletitle{Mat: Mask-aware transformer for large hole image inpainting}. In \bibinfo{booktitle}{\emph{Proceedings of the IEEE/CVF conference on computer vision and pattern recognition}}. \bibinfo{pages}{10758--10768}.
\newblock


\bibitem[Liu et~al\mbox{.}(2019)]%
        {liu2019roberta}
\bibfield{author}{\bibinfo{person}{Yinhan Liu}, \bibinfo{person}{Myle Ott}, \bibinfo{person}{Naman Goyal}, \bibinfo{person}{Jingfei Du}, \bibinfo{person}{Mandar Joshi}, \bibinfo{person}{Danqi Chen}, \bibinfo{person}{Omer Levy}, \bibinfo{person}{Mike Lewis}, \bibinfo{person}{Luke Zettlemoyer}, {and} \bibinfo{person}{Veselin Stoyanov}.} \bibinfo{year}{2019}\natexlab{}.
\newblock \showarticletitle{Roberta: A robustly optimized bert pretraining approach}.
\newblock \bibinfo{journal}{\emph{arXiv:1907.11692}} (\bibinfo{year}{2019}).
\newblock


\bibitem[Ltd.({[n.\,d.]})]%
        {Eyes_Japan}
\bibfield{author}{\bibinfo{person}{Eyes JAPAN~Co. Ltd.}} \bibinfo{year}{[n.\,d.]}\natexlab{}.
\newblock \bibinfo{title}{{Eyes Japan MoCap Dataset}}.
\newblock
\newblock
\urldef\tempurl%
\url{http://mocapdata.com}
\showURL{%
\tempurl}


\bibitem[Mahmood et~al\mbox{.}(2019)]%
        {mahmood2019amass}
\bibfield{author}{\bibinfo{person}{Naureen Mahmood}, \bibinfo{person}{Nima Ghorbani}, \bibinfo{person}{Nikolaus~F Troje}, \bibinfo{person}{Gerard Pons-Moll}, {and} \bibinfo{person}{Michael~J Black}.} \bibinfo{year}{2019}\natexlab{}.
\newblock \showarticletitle{AMASS: Archive of motion capture as surface shapes}. In \bibinfo{booktitle}{\emph{ICCV}}. \bibinfo{pages}{5442--5451}.
\newblock


\bibitem[Mandery et~al\mbox{.}(2016)]%
        {mandery2016unifying}
\bibfield{author}{\bibinfo{person}{Christian Mandery}, \bibinfo{person}{{\"O}mer Terlemez}, \bibinfo{person}{Martin Do}, \bibinfo{person}{Nikolaus Vahrenkamp}, {and} \bibinfo{person}{Tamim Asfour}.} \bibinfo{year}{2016}\natexlab{}.
\newblock \showarticletitle{Unifying representations and large-scale whole-body motion databases for studying human motion}.
\newblock \bibinfo{journal}{\emph{IEEE Transactions on Robotics}} \bibinfo{volume}{32}, \bibinfo{number}{4} (\bibinfo{year}{2016}), \bibinfo{pages}{796--809}.
\newblock


\bibitem[Mandery et~al\mbox{.}(2015)]%
        {7251476}
\bibfield{author}{\bibinfo{person}{Christian Mandery}, \bibinfo{person}{Ömer Terlemez}, \bibinfo{person}{Martin Do}, \bibinfo{person}{Nikolaus Vahrenkamp}, {and} \bibinfo{person}{Tamim Asfour}.} \bibinfo{year}{2015}\natexlab{}.
\newblock \showarticletitle{The KIT whole-body human motion database}. In \bibinfo{booktitle}{\emph{2015 International Conference on Advanced Robotics (ICAR)}}. \bibinfo{pages}{329--336}.
\newblock
\urldef\tempurl%
\url{https://doi.org/10.1109/ICAR.2015.7251476}
\showDOI{\tempurl}


\bibitem[M\"{u}ller et~al\mbox{.}(2007)]%
        {MPI_HDM05}
\bibfield{author}{\bibinfo{person}{M. M\"{u}ller}, \bibinfo{person}{T. R\"{o}der}, \bibinfo{person}{M. Clausen}, \bibinfo{person}{B. Eberhardt}, \bibinfo{person}{B. Kr\"{u}ger}, {and} \bibinfo{person}{A. Weber}.} \bibinfo{year}{2007}\natexlab{}.
\newblock \bibinfo{booktitle}{\emph{Documentation Mocap Database {HDM05}}}.
\newblock \bibinfo{type}{{T}echnical {R}eport} CG-2007-2. \bibinfo{institution}{Universit\"{a}t Bonn}.
\newblock


\bibitem[Pavllo et~al\mbox{.}(2020)]%
        {pavllo2020modeling}
\bibfield{author}{\bibinfo{person}{Dario Pavllo}, \bibinfo{person}{Christoph Feichtenhofer}, \bibinfo{person}{Michael Auli}, {and} \bibinfo{person}{David Grangier}.} \bibinfo{year}{2020}\natexlab{}.
\newblock \showarticletitle{Modeling human motion with quaternion-based neural networks}.
\newblock \bibinfo{journal}{\emph{International Journal of Computer Vision}}  \bibinfo{volume}{128} (\bibinfo{year}{2020}), \bibinfo{pages}{855--872}.
\newblock


\bibitem[Punnakkal et~al\mbox{.}(2021)]%
        {punnakkal2021babel}
\bibfield{author}{\bibinfo{person}{Abhinanda~R Punnakkal}, \bibinfo{person}{Arjun Chandrasekaran}, \bibinfo{person}{Nikos Athanasiou}, \bibinfo{person}{Alejandra Quiros-Ramirez}, {and} \bibinfo{person}{Michael~J Black}.} \bibinfo{year}{2021}\natexlab{}.
\newblock \showarticletitle{BABEL: Bodies, action and behavior with english labels}. In \bibinfo{booktitle}{\emph{Proceedings of the IEEE/CVF Conference on Computer Vision and Pattern Recognition}}. \bibinfo{pages}{722--731}.
\newblock


\bibitem[Qin et~al\mbox{.}(2022)]%
        {qin2022twostage}
\bibfield{author}{\bibinfo{person}{Jia Qin}, \bibinfo{person}{Youyi Zheng}, {and} \bibinfo{person}{Kun Zhou}.} \bibinfo{year}{2022}\natexlab{}.
\newblock \showarticletitle{Motion In-betweening via Two-stage Transformers}.
\newblock \bibinfo{journal}{\emph{ACM TOG}} \bibinfo{volume}{41}, \bibinfo{number}{6} (\bibinfo{year}{2022}), \bibinfo{pages}{1--16}.
\newblock


\bibitem[Sigal et~al\mbox{.}(2010)]%
        {HumanEva}
\bibfield{author}{\bibinfo{person}{L. Sigal}, \bibinfo{person}{A. Balan}, {and} \bibinfo{person}{M.~J. Black}.} \bibinfo{year}{2010}\natexlab{}.
\newblock \showarticletitle{{HumanEva}: Synchronized video and motion capture dataset and baseline algorithm for evaluation of articulated human motion}.
\newblock \bibinfo{journal}{\emph{International Journal of Computer Vision}} \bibinfo{volume}{87}, \bibinfo{number}{4} (\bibinfo{date}{March} \bibinfo{year}{2010}), \bibinfo{pages}{4--27}.
\newblock
\urldef\tempurl%
\url{https://doi.org/10.1007/s11263-009-0273-6}
\showDOI{\tempurl}


\bibitem[Taheri et~al\mbox{.}(2020)]%
        {taheri2020grab}
\bibfield{author}{\bibinfo{person}{Omid Taheri}, \bibinfo{person}{Nima Ghorbani}, \bibinfo{person}{Michael~J Black}, {and} \bibinfo{person}{Dimitrios Tzionas}.} \bibinfo{year}{2020}\natexlab{}.
\newblock \showarticletitle{GRAB: A dataset of whole-body human grasping of objects}. In \bibinfo{booktitle}{\emph{Computer Vision--ECCV 2020: 16th European Conference, Glasgow, UK, August 23--28, 2020, Proceedings, Part IV 16}}. Springer, \bibinfo{pages}{581--600}.
\newblock


\bibitem[Tevet et~al\mbox{.}(2022)]%
        {tevet2022humanmotiondiffusion}
\bibfield{author}{\bibinfo{person}{Guy Tevet}, \bibinfo{person}{Sigal Raab}, \bibinfo{person}{Brian Gordon}, \bibinfo{person}{Yonatan Shafir}, \bibinfo{person}{Daniel Cohen-Or}, {and} \bibinfo{person}{Amit~H Bermano}.} \bibinfo{year}{2022}\natexlab{}.
\newblock \showarticletitle{Human motion diffusion model}.
\newblock \bibinfo{journal}{\emph{arXiv:2209.14916}} (\bibinfo{year}{2022}).
\newblock


\bibitem[Troje(2002)]%
        {BMLrub}
\bibfield{author}{\bibinfo{person}{Nikolaus~F. Troje}.} \bibinfo{year}{2002}\natexlab{}.
\newblock \showarticletitle{Decomposing Biological Motion: {A} Framework for Analysis and Synthesis of Human Gait Patterns}.
\newblock \bibinfo{journal}{\emph{Journal of Vision}} \bibinfo{volume}{2}, \bibinfo{number}{5} (\bibinfo{date}{Sept.} \bibinfo{year}{2002}), \bibinfo{pages}{2--2}.
\newblock
\urldef\tempurl%
\url{https://doi.org/10.1167/2.5.2}
\showDOI{\tempurl}


\bibitem[Trumble et~al\mbox{.}(2017)]%
        {Trumble:BMVC:2017}
\bibfield{author}{\bibinfo{person}{Matt Trumble}, \bibinfo{person}{Andrew Gilbert}, \bibinfo{person}{Charles Malleson}, \bibinfo{person}{Adrian Hilton}, {and} \bibinfo{person}{John Collomosse}.} \bibinfo{year}{2017}\natexlab{}.
\newblock \showarticletitle{Total Capture: 3D Human Pose Estimation Fusing Video and Inertial Sensors}. In \bibinfo{booktitle}{\emph{2017 British Machine Vision Conference (BMVC)}}.
\newblock


\bibitem[Vaswani et~al\mbox{.}(2017)]%
        {vaswani2017attention}
\bibfield{author}{\bibinfo{person}{Ashish Vaswani}, \bibinfo{person}{Noam Shazeer}, \bibinfo{person}{Niki Parmar}, \bibinfo{person}{Jakob Uszkoreit}, \bibinfo{person}{Llion Jones}, \bibinfo{person}{Aidan~N Gomez}, \bibinfo{person}{{\L}ukasz Kaiser}, {and} \bibinfo{person}{Illia Polosukhin}.} \bibinfo{year}{2017}\natexlab{}.
\newblock \showarticletitle{Attention is all you need}.
\newblock \bibinfo{journal}{\emph{Advances in neural information processing systems}}  \bibinfo{volume}{30} (\bibinfo{year}{2017}).
\newblock


\bibitem[Zhang et~al\mbox{.}(2022)]%
        {zhang2022motiondiffuse}
\bibfield{author}{\bibinfo{person}{Mingyuan Zhang}, \bibinfo{person}{Zhongang Cai}, \bibinfo{person}{Liang Pan}, \bibinfo{person}{Fangzhou Hong}, \bibinfo{person}{Xinying Guo}, \bibinfo{person}{Lei Yang}, {and} \bibinfo{person}{Ziwei Liu}.} \bibinfo{year}{2022}\natexlab{}.
\newblock \showarticletitle{Motiondiffuse: Text-driven human motion generation with diffusion model}.
\newblock \bibinfo{journal}{\emph{arXiv:2208.15001}} (\bibinfo{year}{2022}).
\newblock


\bibitem[Zhang et~al\mbox{.}(2023)]%
        {zhang2023remodiffuse}
\bibfield{author}{\bibinfo{person}{Mingyuan Zhang}, \bibinfo{person}{Xinying Guo}, \bibinfo{person}{Liang Pan}, \bibinfo{person}{Zhongang Cai}, \bibinfo{person}{Fangzhou Hong}, \bibinfo{person}{Huirong Li}, \bibinfo{person}{Lei Yang}, {and} \bibinfo{person}{Ziwei Liu}.} \bibinfo{year}{2023}\natexlab{}.
\newblock \showarticletitle{ReMoDiffuse: Retrieval-Augmented Motion Diffusion Model}.
\newblock \bibinfo{journal}{\emph{arXiv preprint arXiv:2304.01116}} (\bibinfo{year}{2023}).
\newblock


\bibitem[Zhang and Tang(2022)]%
        {zhang2022gamma}
\bibfield{author}{\bibinfo{person}{Yan Zhang} {and} \bibinfo{person}{Siyu Tang}.} \bibinfo{year}{2022}\natexlab{}.
\newblock \showarticletitle{The wanderings of odysseus in 3D scenes}. In \bibinfo{booktitle}{\emph{CVPR}}. \bibinfo{pages}{20481--20491}.
\newblock


\bibitem[Zhou et~al\mbox{.}(2019)]%
        {zhou2019continuity}
\bibfield{author}{\bibinfo{person}{Yi Zhou}, \bibinfo{person}{Connelly Barnes}, \bibinfo{person}{Jingwan Lu}, \bibinfo{person}{Jimei Yang}, {and} \bibinfo{person}{Hao Li}.} \bibinfo{year}{2019}\natexlab{}.
\newblock \showarticletitle{On the continuity of rotation representations in neural networks}. In \bibinfo{booktitle}{\emph{Proceedings of the IEEE/CVF Conference on Computer Vision and Pattern Recognition}}. \bibinfo{pages}{5745--5753}.
\newblock


\end{thebibliography}
\appendix

\section{Details of Progressive Mask Transformer}
In this section, we introduce the details of the proposed progressive mask transformer. We adopt the similar motion representation, position encoding, and loss functions introduced in two-stage transformers \cite{qin2022twostage} to evaluate the proposed attention mechanism and progressive generation strategy.

\noindent\textbf{Motion Representation}. We follow the same representation used in \cite{qin2022twostage} and adopt the continuous 6D rotation representation introduced by \cite{zhou2019continuity}. 
Each frame is represented as $x_t=\{r_t, p_t, c_t, m_t\}$, where $r_t \in \mathbb{R}^{J \times 6}$ is the first two rows or columns of $3 \times 3$ rotation matrix
and $p_t \in \mathbb{R}^{3}$ is the world position of the hip joint.
%For the deep-learning methods, other features are included. $c_t \in \{0, 1\}^4$ represents whether the feet have contact with the ground and $m_t \in \{0, 1\}$ represents whether this frame is masked. Finally the motion clip $x \in \mathbb{R}^{T \times D}$, where $D=J\times6+8$ for deep-learning methods and $D=J\times6+3$ for traditional methods. 
$c_t \in \{0, 1\}^4$ represents whether the feet have contact with the ground and $m_t \in \{0, 1\}$ represents whether this frame is masked. Finally the motion clip $x \in \mathbb{R}^{T \times D}$, where $D=J\times6+8$ and $T$ is the number of frames.

\noindent\textbf{Progressive Generation.}
Rather than employing a two-stage approach with separate models for coarse and fine generation \cite{qin2022twostage}, inspired by the progressive strategy in other generation tasks~\cite{maskpredict}, we utilize a single network that refines its output iteratively.
The proposed progressive generation strategy runs for a predetermined number $r$ (e.g., 3) of iterations to gradually recover the masked motion. It takes a masked motion $m_x \in \mathbb{R}^{T \times D}$ as input and generates a motion $m_{xi} \in \mathbb{R}^{T \times D}$, where $i$ is the iteration number.
The output of each iteration will be the input of the next iteration, and the final motion is obtained after $r$ iterations.
%We use a Markov chain generation process, but all tokens will remain in every iteration. 
%The two-stage transformers use two models for coarse generation and fine generation, which can be seen as an assembled learning method. We instead use one network and let it refine its output itself, inspired by the progressive strategy used in other generation tasks~\cite{maskpredict}.
%, we let the transformer's attention mechanism choose the tokens it wants. 
Using the progressive generation strategy, we find that the network first generates coarse results in the first iteration, which typically contain artifacts like foot sliding and jittering, and then refine it in the following iterations. This approach is both parameter-efficient and can leverage knowledge from a previously well-learned model.

\noindent\textbf{Dynamic Attention Mask.}
A notable observation of previous works \cite{qin2022twostage} \cite{li2022mat} is that for motions dominated by mission frames, the vanilla transformer's default attention strategy may undermine the valid information, which causes limited generalization ability. And thus the masked multi-head self-attention is used to mask out missing frames. However, this strategy may not be optimal for motion blending since it cuts off the connection between most of the tokens. Inspired by the technique in image in-painting \cite{li2022mat}, we use the updated attention mask instead. We proposed a simple yet efficient masking strategy that first masks all missing frames similar to \cite{qin2022twostage} but later gradually reduces the number of masked frames. This strategy is naturally combined with the aforementioned progressive generation method. Different attention masks will be used in different iterations of the generation process. At the first iteration, all missing frames are masked. In the following iteration, the number of masked frames will linearly decrease.

\noindent\textbf{Position Encoding}. The position encoding contains two parts: the keyframe position encoding and the step embedding. The keyframe positional encoding \cite{qin2022twostage} uses an MLP with a single hidden layer of 512 units to encode the relative position from the current frame to keyframes, which is added to the motion embedding and fed to the transformer backbone. Incorporating the progressive generation strategy, we introduce the concept of learned step embedding, which assigns a distinct vector to each iteration step, allowing the model to differentiate between different iteration time points.
The step embedding is similar to the relative positional encoding proposed in the two-stage transformers \cite{qin2022twostage}, which is a learnable lookup table containing $2 \times T - 1$ embeddings, serving as shared key tokens for calculating attention in all transformer layers.

\iffalse
\begin{figure*}[hbt!]
  \centering
  \vspace{-3mm}
  \includegraphics[width=0.9\linewidth]{images/Figure3.transformer.pdf}
  \vspace{-3mm}
  \caption{Our method uses a progressive generation strategy, which iteratively generates a motion with an updated attention mask and different relative position encoding. In each iteration, a refined motion with more details and fewer artifacts is generated. The trained network could also act as a neural post-processor that alleviates foot sliding and jittering.}
  \Description{The overview of the method.}
  \label{fig:transformer}
\end{figure*}
\fi

\noindent\textbf{Loss Functions}. In order to fairly compare different methods, for the experiments on the AMASS dataset we train all models using the same loss functions as follows.

\noindent\emph{State Loss}.
The state of motion contains local rotation in 6D rotation space, root position, and foot contact. The state loss is a weighted sum of the reconstruction loss of these three components:
\begin{equation} 
L_{state}=
\lambda_c L1(c, \hat{c})+
\lambda_r \hat{L1}(r, \hat{r})+
\lambda_p \hat{L1}(p, \hat{p}),
\end{equation}
where $L1$ is the mean absolute error, and $\hat{L1}$ stands for smooth L1 loss which uses a squared term if the absolute element-wise error falls below a threshold and an L1 term otherwise.

\noindent\emph{Joint Position Loss}. Considering equally distributed joint orientation errors will lead to growing joint position
errors along the kinematic chains, the reconstruction of joint positions is commonly used in motion generation tasks \cite{pavllo2020modeling}. Besides, the smoothness term of joint positions is also included: 
\begin{equation} 
L_{pos}=
\hat{L1}(g, \hat{g}) + \lambda_s||\hat{g}'||_1,
\end{equation}
where $g$ stands for the global joint position, which is calculated by root position and local joint rotations through forward kinematics, and $\hat{g}'$ is the speed of joint positions.

\noindent\emph{Foot Sliding Loss}. This term is calculated based on the predicted foot contact and the generated joint position speed. When foot contact happens, the speed of foot joints should be zero:
\begin{equation} 
L_{foot}=||\hat{c}\hat{g}_{foot}'||_1,
\end{equation}

\begin{table*}[t] 
    %\vspace{-1mm}
    \caption{Motion Blending Ablation, all models are trained with random transition lengths from 5 to 60. The proposed Dynamic Attention Mask (DAM) and Progressive Generation (PG) offer promising enhancements to the strong baseline method \cite{qin2022twostage}.}
    \label{tab:amassExp}
    %\vspace{-3mm}
    \centering
    \resizebox{\textwidth}{!}{
    \begin{tabular}{lllllllllllllll}
    \toprule
        ~ & Param & \multicolumn{6}{c}{Pos / m} & \multicolumn{6}{c}{Rotation} & Mean \\ 
        Frames & ~ & 5 & 15 & 30 & 45 & 60 & 70 & 5 & 15 & 30 & 45 & 60 & 70  \\ 
    \midrule
Detail Trans \cite{qin2022twostage}&20.52M&\color{blue}{0.142} &\color{blue}{0.398} &0.762 &1.082 &1.358 &1.576 &\color{blue}{0.123} &\color{blue}{0.265} &\color{blue}{0.465} &\color{blue}{0.624} &0.744 &0.828 &0.697 \\
ours w/o PG and DAM \cite{qin2022twostage}&10.39M&0.243 &0.488 &0.829 &1.145 &1.416 &1.637 &0.195 &0.321 &0.509 &0.662 &0.778 &0.863 &0.757 \\
ours w/o PG &10.39M&0.224 &0.459 &0.806 &1.129 &1.396 &1.630 &0.177 &0.305 &0.499 &0.654 &0.762 &0.846 &0.741 \\
ours&10.39M&0.167 &0.402 &\color{blue}{0.728} &\color{blue}{1.030} &\color{blue}{1.296} &\color{red}{1.506} &0.161 &0.290 &0.474 &0.636 &\color{blue}{0.735} &\color{blue}{0.819} &\color{blue}{0.687} \\
ours (two stages)&20.77M&\color{red}{0.122} &\color{red}{0.379} &\color{red}{0.714} &\color{red}{1.022} &\color{red}{1.291} &\color{blue}{1.508} &\color{red}{0.113} &\color{red}{0.262} &\color{red}{0.455} &\color{red}{0.621} &\color{red}{0.726} &\color{red}{0.817} &\color{red}{0.669} \\
    \bottomrule
    \end{tabular}
}
%\vspace{-2mm}
\end{table*}

\section{Details of Experiments}
Our experiments are conducted on the AMASS database, known for its popularity in motion generation. We use different scales of data in different sub-tasks for fair comparisons with previous methods. The metrics include L2P, L2Q, physics error, and trajectory error. The L2P and L2Q measure the average L2 distance of the global joint position and rotation (in quaternions) per joint per frame. The physics error is the sum of the foot floating distance and the foot-ground penetration distance. The trajectory error measures the average error between the desired path and the hip position of the character per frame.

\noindent\textbf{Trajectory Following}. 
For the trajectory following task, we carefully follow the experiment settings of the GAMMA \cite{zhang2022gamma} and use the same randomly chosen input seed poses and data in our experiments. The training data contains CMU \cite{CMU}, MPI HDM05 \cite{MPI_HDM05}, BMLmovi \cite{ghorbani2021movi}, KIT \cite{mandery2016unifying}, Eyes Japan \cite{Eyes_Japan}. The evaluation data includes HumanEva \cite{HumanEva}, and ACCAD \cite{ACCAD}. 
In practical applications, it’s not only necessary to achieve high trajectory matching but also to reach the endpoint within a specified time. Therefore, we propose a metric that includes velocity: we defined trajectory as position-time pairs and used the point-wise L2 distance error. We mainly focus on this metric although FID or foot sliding will also be helpful, considering our method is retrieval-based and the motion quality is rather high. 
We follow previous work \cite{holden2017phase} and use different trajectories for evaluation. For the wave trajectory, we define it as a sine function about time $x(t) = 2 sin(t)$. For the square trajectory, we set the side length as 5. For the circle trajectory, the diameter is set as 5. We randomly select 50 seed poses from HumanEva and ACCAD respectively. Both methods generate a 20-22.5-second motion based on each seed pose.

\noindent\textbf{Temporal Action Composition}. 
For temporal action composition, the label we use is coarse and noisy but simple to collect. Note that TEACH uses the BABEL dataset \cite{punnakkal2021babel} which is a subset of AMASS that contains detailed text descriptions, in total about 8000 motion names and 43 hours of motion data. 
We select all motions in AMASS that have meaningful file names and are present in BABEL for a fair comparison with TEACH. In total, we collate 60 motion names such as ``walk fast'', ``turn left'', ``wave'' and ``air guitar''.
We ask the language model (ChatGPT \cite{chatGPT}) to choose 5 motions from these motion names and create a short story about a human doing some activities. The evaluated methods are then asked to generate a 12.5-second motion based on these 5 motion names.

\noindent\textbf{Motion Blending}. 
Motion blending can be viewed as motion completion \cite{duan2021single}, namely, a number of the motion frames in the center temporal position will be masked, and the goal is to recover the original motion. We extensively evaluate models on transitions spanning from 5 to 70 frames, which is challenging because AMASS contains motions consisting of sporadic, random short movements that are extremely difficult to predict beyond short time horizons \cite{harvey2020robust}. In addition to the data employed for the trajectory following task, GRAB \cite{taheri2020grab}, DanceDB \cite{DanceDB}, BMLrub \cite{BMLrub}, and WEIZMANN \cite{7251476} are added to training data, in total 402 subjects, 15818 motions, and 56.3 hours motion data. %107 minutes of motion data performed by 32 objects covering 666 motion types are used for evaluation, including HumanEva, ACCAD, EKUT \cite{KrebsMeixner2021}, and TotalCapture \cite{Trumble:BMVC:2017}. 
For evaluation, we utilize 107 minutes of motion data performed by 32 objects, encompassing 666 motions, which includes contributions from various sources: HumanEva, ACCAD, EKUT \cite{KrebsMeixner2021}, and TotalCapture \cite{Trumble:BMVC:2017}.

\emph{Ablation Study}. In Table \ref{tab:amassExp}, we also evaluate the impact of different components on the final performance. The Context Transformer (row 2) \cite{qin2022twostage} is our baseline. Row 3 is our method without the progressive generation strategy, which simply replaces the Full Attention Mask used in the Context Transformer as the proposed Dynamic Attention Mask. The comparison of row 2 and row 3 shows the efficiency of the Dynamic Attention Mask. The full method (row 4) utilizes the proposed progressive generation strategy, which brings a further improvement compared with using Dynamic Attention Mask only (row 3). In a manner akin to the two-stage approach \cite{qin2022twostage}, our method demonstrates additional advancements, leading to a notable enhancement (rows 4 and 5). Our method also surpasses the state-of-the-art (row 1).%In comparison to the state-of-the-art method (row 1), the proposed method showcases a distinct advantage.

\begin{figure*}[t]
  \centering
  \includegraphics[width=\linewidth]{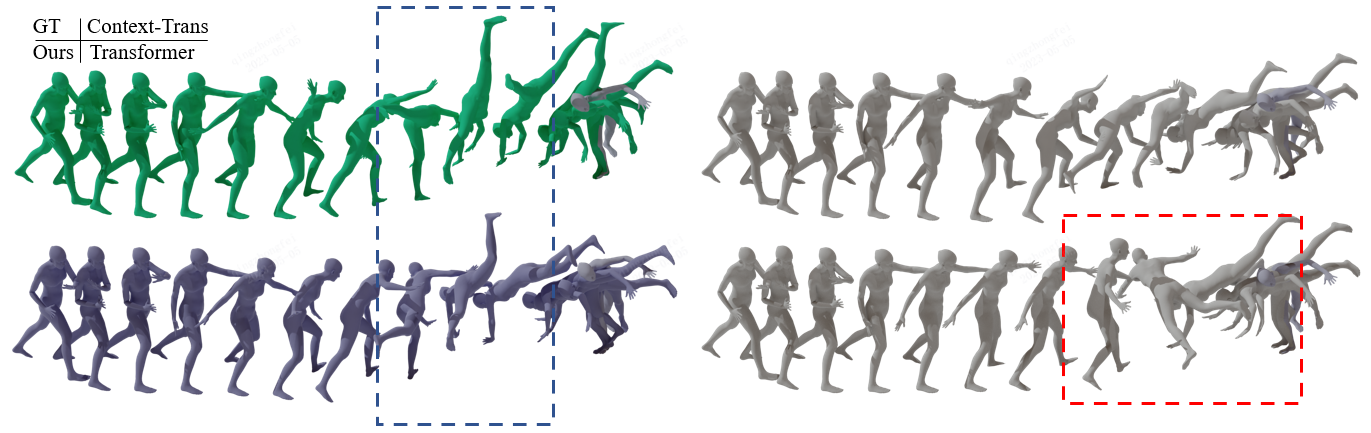}
  \caption{Smoothness and detail comparison. For complex motion, the proposed method generates smooth motion while maintaining high-quality details. Note that in this image the position of the actor is changed (using the same way) for better visibility.}
  \label{fig:flip}
  \Description{flip motion comparison.}

\end{figure*}

\emph{Qualitative Comparison}.
In this section, we show a qualitative comparison of the evaluation dataset. We compare our method (shown in blue) with the Context Transformer (shown in grey and upper) and Vanilla Transformer (shown in grey) \cite{vaswani2017attention} together with the ground truth (shown in green). As shown in Figure \ref{fig:flip}, the Vanilla Transformer may generate motions with inconsistent speed. Notably, our method can generate better high-frequency details while maintaining a consistent speed. When generating long motion, foot sliding is a longstanding problem. The Context Transformer and Vanilla Transformer cause noticeable foot sliding (see the leg foot in the red box). The proposed method generates motion with high-frequency details (more footsteps which is similar to the ground truth) while alleviating the foot sliding.

\begin{figure*}[t]
  \centering
  \includegraphics[width=\linewidth]{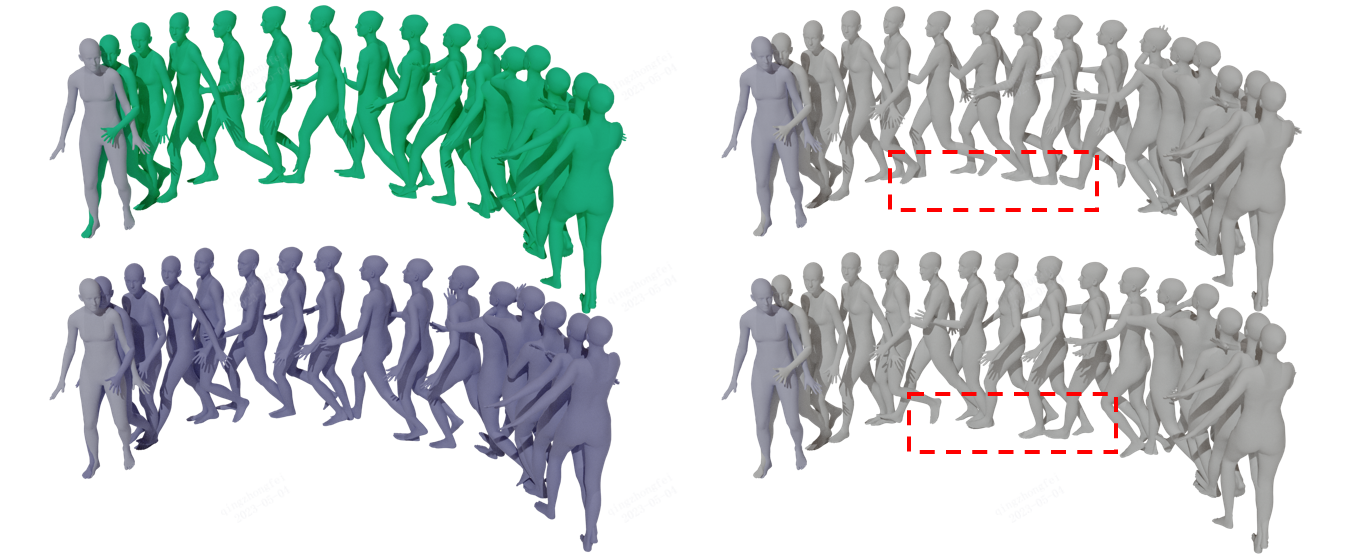}
  \caption{Foot sliding comparison. Our method generates motions with more high-frequency details while alleviating foot sliding artifacts.}
  \label{fig:foot}
  \Description{foot sliding.}

\end{figure*}
\end{document}